\documentclass[runningheads,orivec]{llncs}
\usepackage[T1]{fontenc}
\usepackage{graphicx}
\usepackage{orcidlink}
\usepackage[acronym]{glossaries}
\usepackage{subfigure}
\usepackage{multirow}
\usepackage[misc,geometry]{ifsym}

\newacronym{iac}{IAC}{intracranial arterial calcification}
\newacronym{ct}{CT}{computer tomography}
\newacronym{ssim}{SSIM}{structural similarity index measure}
\newacronym{ist3}{IST-3}{third International Stroke Trial}
\newacronym[plural=RoIs, longplural=regions of interest]{roi}{RoI}{region of interest}
\newacronym{mri}{MRI}{magnetic resonance imaging}
\newacronym{mae}{MAE}{masked autoencoder}
\newacronym{vit}{ViT}{Vision Transformer}

\begin{document}
\title{Calibrated Self-supervised Vision Transformers Improve Intracranial Arterial Calcification Segmentation from Clinical CT Head Scans}
\titlerunning{Calibrated Vision Transformers Improve IAC Segmentation}
%
\author{Benjamin Jin\inst{1}\textsuperscript{(\Letter)}\orcidlink{0009-0006-2710-7248} \and
Grant Mair\inst{1,2}\orcidlink{0000-0003-2189-443X} \and
Joanna M. Wardlaw\inst{1,3}\orcidlink{0000-0002-9812-6642} \and
Maria del C. Valdés Hernández\inst{1,3}\orcidlink{0000-0003-2771-6546}}
\authorrunning{B. Jin et al.}
%

\institute{
Centre for Clinical Brain Sciences, University of Edinburgh,  Edinburgh, UK\\
    \email{b.jin@ed.ac.uk}\and
Neuroradiology, Department of Clinical Neurosciences, NHS Lothian, Edinburgh, UK \and
UK Dementia Research Institute, University of Edinburgh, Edinburgh, UK
}%
\maketitle              
\setcounter{footnote}{0} 
\begin{abstract}

\Acrfullpl{vit} have gained significant popularity in the natural image domain but have been less successful in 3D medical image segmentation. Nevertheless, 3D \acrshortpl{vit} are particularly interesting for large medical imaging volumes due to their efficient self-supervised training within the \acrfull{mae} framework, which enables the use of imaging data without the need for expensive manual annotations. \Acrfull{iac} is an imaging biomarker visible on routinely acquired CT scans linked to neurovascular diseases such as stroke and dementia, and automated \acrshort{iac} quantification could enable their large-scale risk assessment. We pre-train \acrshortpl{vit} with \acrshort{mae} and fine-tune them for \acrshort{iac} segmentation for the first time. To develop our models, we use highly heterogeneous data from a large clinical trial, the \acrfull{ist3}. We evaluate key aspects of \acrshort{mae} pre-trained \acrshortpl{vit} in \acrshort{iac} segmentation, and analyse the clinical implications. We show: 1) our calibrated self-supervised \acrshort{vit} beats a strong supervised nnU-Net baseline by 3.2 Dice points, 2) low patch sizes are crucial for \acrshortpl{vit} for \acrshort{iac} segmentation and interpolation upsampling with regular convolutions is preferable to transposed convolutions for \acrshort{vit}-based models, and 3) our \acrshortpl{vit} increase robustness to higher slice thicknesses and improve risk group classification in a clinical scenario by 46\%. Our code is available online\footnote{Code at \url{https://github.com/bjin96/mae-medical-segmentation/}}.

\end{abstract}

\section{Introduction}

\Acrfull{iac} is a biomarker of intracranial atherosclerosis, a disease in which the head arteries become damaged over time \cite{Libby19}, leading to an increased risk of neurovascular diseases such as stroke \cite{Banerjee17} and dementia \cite{Chen19}. Due to its high density, \acrshort{iac} is visible on routinely acquired CT scans. In Scotland alone, a country with a population of 5.5 million people, more than 100,000 CT scans are collected every year (across all examined body parts) \cite{Baxter24}, providing a rich resource. Manual \acrshort{iac} detection, even with efficient visual scoring methods \cite{Subedi15}, is not feasible at this scale -- fully automated methods are necessary. Automated \acrshort{iac} detection could aid individualised risk assessment and intervention in routine clinical practice, and support recruitment for research studies into treatments for patients with heightened risk of neurovascular diseases.

We propose an \acrshort{iac} segmentation method using 3D \acrfullpl{vit} and self-supervised pre-training. Originating in work for natural language processing, transformer models have subsequently been adapted for natural images \cite{Dosovitskiy21}. Combining low inductive biases (architectural design choices to facilitate the recognition of specific patterns) with excellent scaling behaviours and large-scale training with huge datasets has led to massive success in both domains and beyond \cite{Islam23}. Part of the success of transformer models is the elegant formulation of a self-supervised pre-training paradigm, where the model predicts missing parts of an input sequence to learn the general data structure. The pre-training generally increases model performance when fine-tuned for downstream tasks compared to randomly initialised models \cite{zhang22}. With medical computer vision moving towards foundation models, we argue that \acrshortpl{vit} for medical image segmentation cannot be overlooked. Sparse training within the \acrfull{mae} framework reduces the computational resources necessary for pre-training and decreases the dependency on custom data augmentation \cite{He21} compared to contrastive learning approaches. We predict fine-tuning decoders on top of \acrshort{mae} pre-trained \acrshortpl{vit} will become increasingly important to adapt foundation models for specific use cases.

Previous work for automated \acrshort{iac} segmentation trained supervised convolutional neural networks \cite{Alajaji23,Bortsova17,Bortsova21} on research-level data with only a few CT scanner models and parameter configurations included. In contrast, we develop our models on highly heterogeneous clinical CT head scans derived from the \acrfull{ist3} \cite{ist3_group12,ist3_group15}, a more challenging but arguably more representative dataset for the demands of real-world applications to clinical data. 

Our contributions are: 1) we propose the first \acrshort{iac} segmentation model using \acrshortpl{vit} and self-supervised pre-training on a clinical dataset, beating a strong nnU-Net baseline, 2) we conduct ablations of key parameters of the model, and 3) we explore the clinical implications of the improved \acrshort{iac} segmentation.

\section{Method}

The training of our models consists of two stages. 1) We pre-train a 3D \acrshort{vit} as an encoder within the \acrshort{mae} framework \cite{He21} on 4933 3D \acrfullpl{roi} around the cavernous sinus of the internal carotid artery. 2) We fine-tune the pre-trained \acrshort{vit} in a supervised setting with an \acrshort{iac} annotated subset of 158 \acrshortpl{roi} used for pre-training. Our code is available (link on page 1).

\subsection{Data}

\begin{figure}[ht!]
    \centering
    \includegraphics[width=0.8\textwidth]{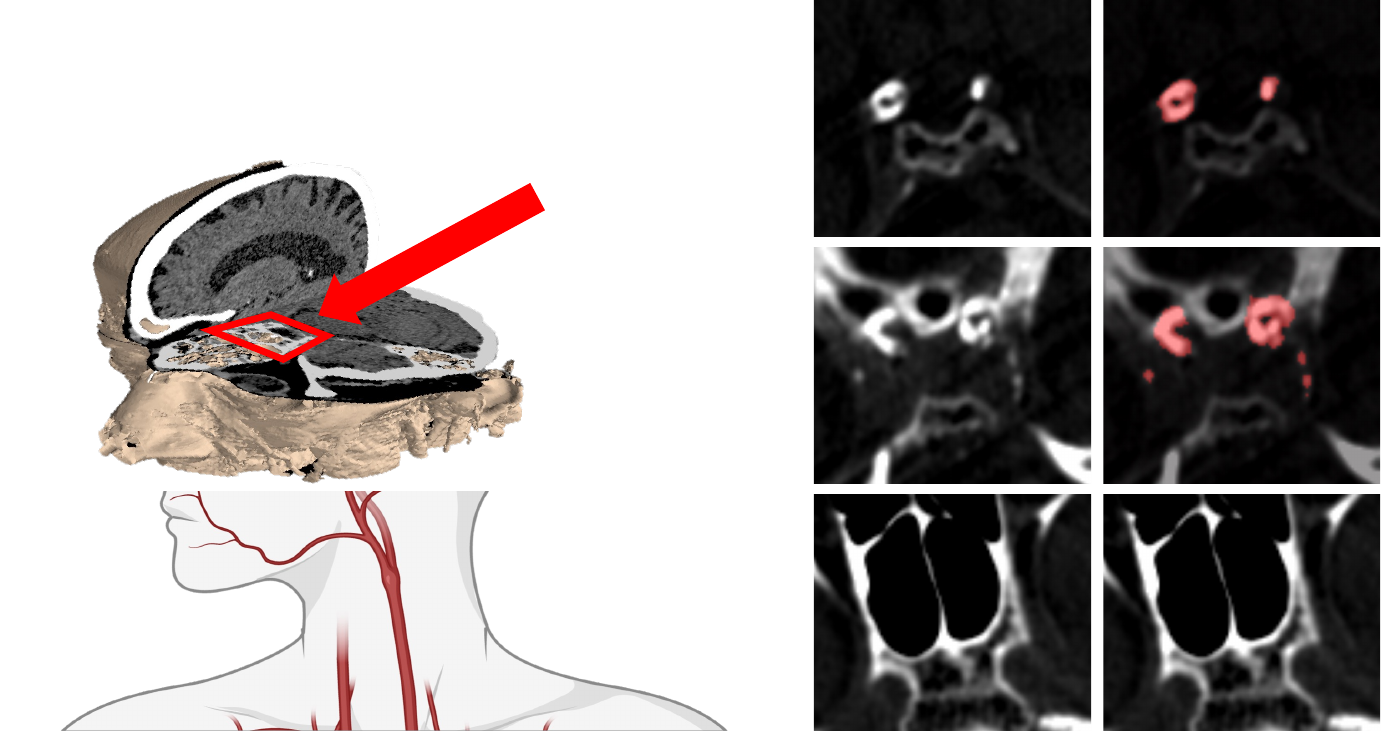}
    \caption{RoI location around the cavernous sinus of the internal carotid artery (left) and selected slices from the RoI (right; lower slices are closer to the neck, upper slices are closer to the skull vault). Intracranial arterial calcification in \textcolor{red}{red} in the right column. Schematic head created in BioRender, \url{BioRender.com/b4yd3yp}.}
    \label{figure:iac_example}
\end{figure}

We use non-enhanced CT scans from the \acrshort{ist3} \cite{ist3_group12,ist3_group15}, a large multicentre randomised controlled study that recruited 3035 patients with acute ischaemic stroke from 156 centres in 12 countries, who gave their informed consent to participate. The CT scans were collected centrally in an anonymised format and assessed visually by expert neuroradiologists as part of the original trial analyses. Here, a large subset of 10,659 CT imaging series from 2,578 patients was available (due to a restriction to non-enhanced CT scans and data corruption).

We previously described the pre-processing and quality control for \acrshort{iac} segmentation with deep learning \cite{jin25}. In brief, CT scans were registered to age-specific templates and cropped to 3D \acrshortpl{roi} around major intracranial arteries defined on the respective templates \cite{Jin2024}. Afterwards, the \acrshortpl{roi} were semi-automatically controlled for successful registration using computational similarity metrics and triaged visual assessment to ensure the crops contained the relevant anatomy. After the pipeline, 6,337 image series from 2,351 patients remained. The image series exhibit extensive heterogeneity with 63 unique CT scanner models from five manufacturers (Siemens, GE, Toshiba, Philips, and Hitachi), the slice thickness ranging from 0.25 to 10 mm, and post-processing by 64 convolution kernels (which are classified into soft tissue and bone kernels). See supplementary Figure 1 for their distributions in the annotated dataset splits.

For the annotations, we follow established practice \cite{Agatston90}: we circle the calcification in 2D, threshold the resulting area to $\geq$ 130 Hounsfield Units, and discard single annotated pixels to avoid segmenting noise. In this work, we only consider the region around the cavernous sinus of the internal carotid artery (\autoref{figure:iac_example}).

\begin{table}
    \centering
    \begin{tabular}{ll|r|r|r}
        \multicolumn{2}{l|}{\textbf{Split}} & \textbf{No. patients} & \textbf{No. image series} & \textbf{With IAC} \\
        \hline
        \multirow{2}{*}{Train (80\%)}
            & Pre-train (100\%) & 1880  & 4933 & n/a \\
            & Fine-tune (10\%) & 188  & 188 & 158 \\
        \multicolumn{2}{l|}{Development (10\%)} & 235  & 235 & 211 \\
        \multicolumn{2}{l|}{Test (10\%)} & 236 & 236 & 209 \\
        \hline
        \multicolumn{2}{l|}{\textbf{Total}} & \textbf{2,351} & \textbf{5,404} & n/a
    \end{tabular}
    \caption{Overview of the dataset splits. Fine-tune data is a subset of the pre-train set.}
    \label{tab:dataset_overview}
\end{table}

We split the data by patients to avoid information leakage from having image series from the same patient in our training and testing datasets (\autoref{tab:dataset_overview}). The development and test sets contain annotated image series from unique patients. Additional image series from these patients are not used (and not all image series contain the region around the cavernous sinus of the internal carotid artery), lowering the total number of imaging series to 5,404 \acrshortpl{roi}. In the random sampling process, we ensure a similar distribution of CT scanner manufacturers as a proxy for similar models and configurations. Annotated image series without calcification present are discarded to avoid undefined overlap metrics.

\subsection{MAE Pre-training}

\begin{figure}
    \centering
    \includegraphics[width=0.66\textwidth]{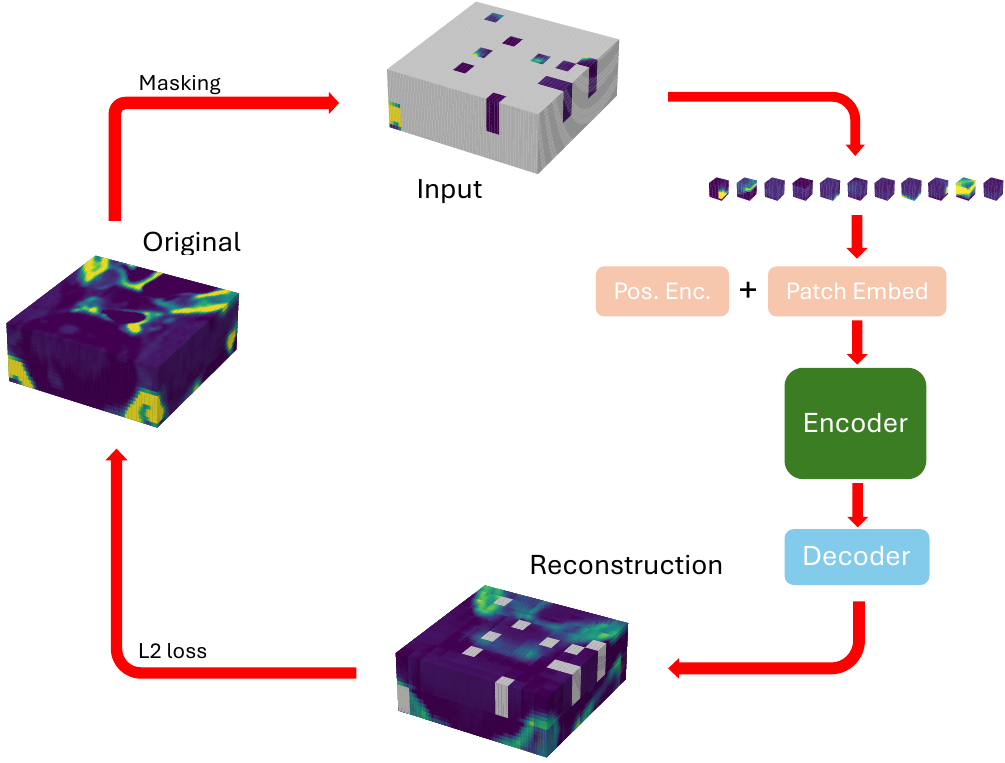}
    \caption{Masked Autoencoder pre-training for IAC segmentation.}
    \label{figure:iac_mae_overview}
\end{figure}

We adapt the original \acrshort{mae} pre-training protocol \cite{He21} for 3D (see \autoref{figure:iac_mae_overview}). First, we project patches (or rather cubes) as tokens into an embedding space, add absolute sin-cos positional encoding extended to encode 3D positions, remove masked tokens, and finally process the tokens with a transformer encoder to output a latent representation. Afterwards, we reconstruct the image series from the latent representation concatenated with mask tokens using a lightweight decoder and compare the reconstruction to the original image series. We experiment with three \acrshort{vit} configurations: ViTiac-S, ViTiac-M, and ViTiac-L (supplementary Table 1), and patch sizes lower than the standard ${16^3}$ to increase the resolution at which the \acrshort{vit} processes the data to accommodate dense prediction.

We increase the percentage of masked patches from 75\% recommended for 2D \cite{He21} to 90\% as recommended for videos \cite{Feichtenhofer22}, since 3D volumes contain more redundant information. We use pixel normalisation before calculating the reconstruction loss and pre-train models for 800 epochs.

\subsection{Supervised Fine-tuning}

\begin{figure}
    \centering
    \includegraphics[width=0.9\textwidth]{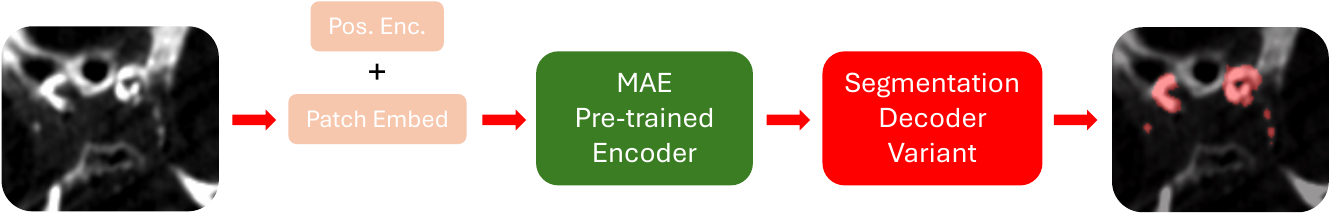}
    \caption{Pre-trained \acrshort{vit} fine-tuning with segmentation decoder variants.}
    \label{figure:fine_tune_overview}
\end{figure}

We fine-tune the pre-trained \acrshort{vit} by adding four different segmentation decoders on top of the pre-trained encoder (\autoref{figure:fine_tune_overview}). We compare three convolutional decoders 1) \textbf{UNETR decoder}: a convolutional decoder ingesting the input volume and outputs from several stages from the transformer encoder \cite{Hatamizadeh21}, 2) \textbf{SFPN U-Net decoder}: a simple feature pyramid \cite{Li22} with a U-Net \cite{Ronneberger15} style decoder, and 3) \textbf{Upscale decoder}: a simple upscaling from the \acrshort{vit} output, and one transformer-based decoder 4) \textbf{MAE decoder}: the same transformer decoder used during \acrshort{mae} pre-training with reset weights. 1) and 2) mainly differ in that the UNETR decoder incorporates the output from the \acrshort{vit}, hidden states from successive stages of the \acrshort{vit}, and the input image series. In contrast, the SFPN U-Net constructs a multi-scale feature pyramid from the final output of the \acrshort{vit} (shown to be preferable for \acrshort{vit} applications \cite{Li22}).

\subsection{Upscaling}

Transposed convolutions are commonly employed in U-Nets for medical image segmentation, but they are known to create checkerboard artefacts \cite{Odena16}. Arguably, checkerboard artefacts are less problematic in U-Nets with convolutional encoders and decoders, as feature maps in every upscaling step are combined with encoder feature maps via skip connections, which help mitigate these artefacts. Models with \acrshort{vit} encoders depend more heavily on upscaling from the internal resolution defined by the patch size. We replace the transposed convolutions (kernel size = $2^3$, stride = $2^3$) in the Upscale decoder with upsampling layers that use nearest-neighbour interpolation followed by a regular convolution (kernel size = $3^3$, stride = $1^3$) to assess the impact of checkerboard artefacts.

\section{Results}

\begin{table}
\setlength{\tabcolsep}{0.3em} 
\renewcommand{\arraystretch}{1.1}
\centering
\begin{tabular}{l|r|r|r|r}
    \textbf{Model} & \textbf{Training} & \textbf{Dice} & \textbf{Volume $\Delta$ [mm$^3$]} \\
    \hline
        nnU-Net Baseline & Superv. & 62.1 [58.7;64.7] &  188[158;225] \\
        Calibrated ViTiac-S & Superv. & 37.1 [34.0;40.6] &  280[253;314] \\
        Calibrated ViTiac-S & SSL+Superv. & \textbf{65.1[62.0;68.4]} & \textbf{100[82;130]} \\ 
    \hline
\end{tabular}
\caption{Mean Dice↑ and mean absolute difference between the predicted and annotated IAC volume denoted as Volume $\Delta$↓ (with 95\% CI from bootstrapping with 1000 resamples) for the supervised nnU-Net Baseline and two ViTiac-S models (with and without MAE pre-training) with the MAE decoder and patch size $4^3$.}
\label{tab:baseline_scratch_comparison}
\end{table}

\begin{figure}[ht]
    \centering
    \includegraphics[width=0.8\textwidth]{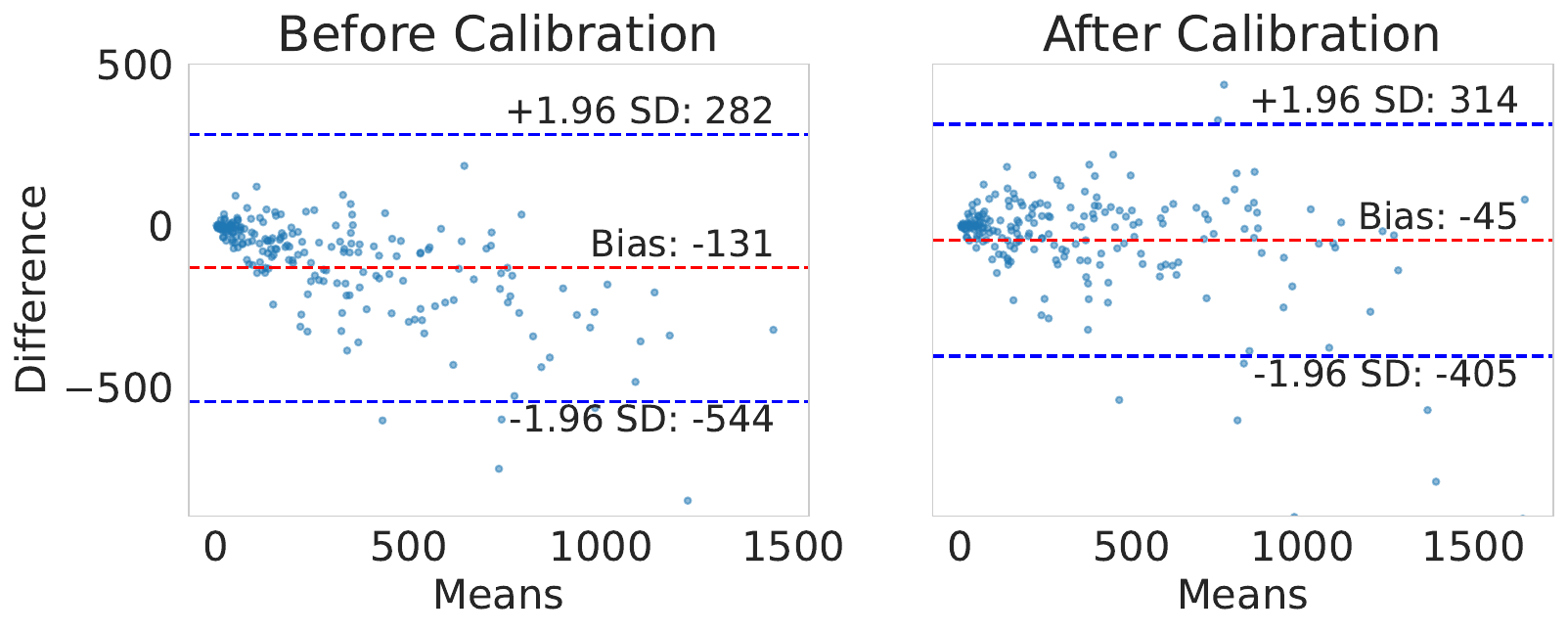}\label{subfig:before_calibration}
    \caption{Bland-Altman plots for the IAC volume (in mm$^3$) predicted by the pre-trained ViTiac-S model compared to the manual annotation before and after calibration.}\label{figure:calibration_plot}
\end{figure}

We calculate the mean Dice and the absolute difference between the predicted and annotated IAC volume to compare a supervised nnU-Net baseline, training a ViTiac-S with MAE decoder from scratch, and pre-training the latter with the MAE framework (\autoref{tab:baseline_scratch_comparison}). The MAE pre-trained ViTiac-S demonstrates the best performance with a Dice score of 65.1 (95\% CI [62.0; 68.4]). We \textit{calibrated} the ViTiac-S models by choosing the decision threshold to minimise the volume difference on the development set (\autoref{figure:calibration_plot}) for the comparison. The pre-trained calibrated ViTiac-S model shows increased robustness for larger slice thicknesses and excels at segmenting medium to large \acrshortpl{iac} compared to the nnU-Net (supplementary Figure 2). While the Dice score is lower than the 76.2 previously reported \cite{Bortsova17}, we again emphasise the heterogeneous clinical nature of our dataset compared to previous research-level data.

\subsection{Ablations}

\begin{figure}
    \centering
    \includegraphics[width=0.99\textwidth]{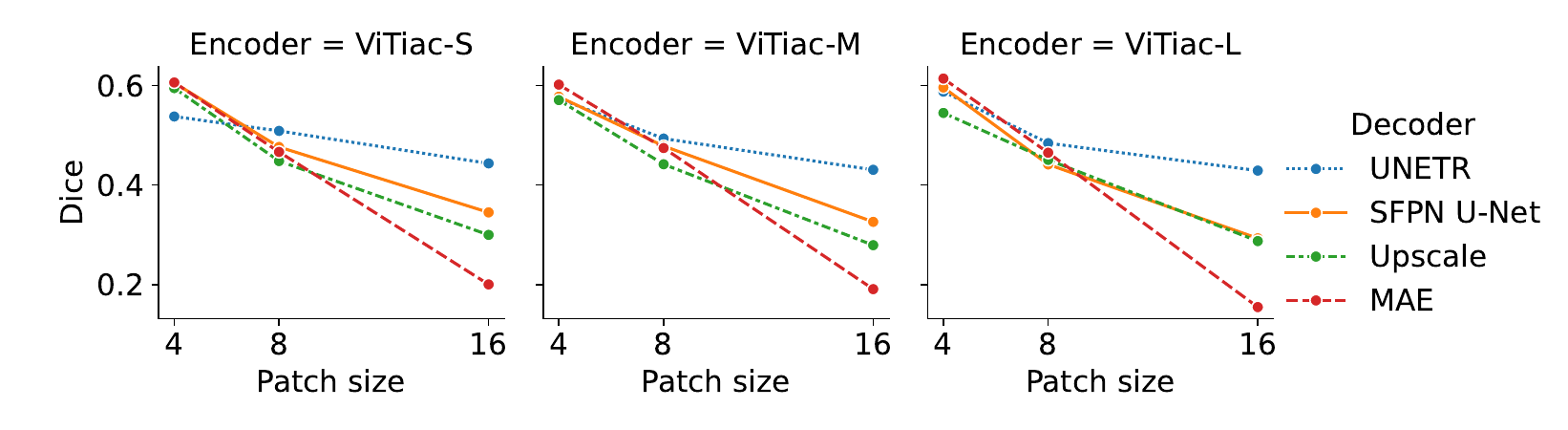}\label{subfig:decoder_comparison_patch_size_dice}
    \caption{Mean Dice↑ of the segmentation decoders under varying patch sizes and MAE pre-trained encoder sizes. }\label{figure:decoder_comparison}
\end{figure}

We compare different segmentation decoders on top of the pre-trained \acrshort{vit} versions and find a large dependence of our models on the patch size. A decrease in patch size (i.e. increase in spatial resolution for the \acrshort{vit}) leads to an increase in Dice score for each \acrshort{mae} pre-trained encoder (ViTiac-x) and every tested segmentation decoder (\autoref{figure:decoder_comparison}). The UNETR decoder degrades the least with increasing patch size. In addition to feature maps from the transformer encoder, the input image series is fed to the convolutional UNETR decoder \textit{directly} and enables the decoder to enrich the low-resolution information from the transformer encoder. The SFPN U-Net and Upscale decoder only ingest the latent output of the transformer encoder, which was previously found to lead to degraded medical image segmentation \cite{Hatamizadeh21}. Our experiments suggest that this is due to the high patch size in their work. The \acrshort{mae} decoder contains no convolutional layers. This results in a juxtaposition of poor segmentation performance for larger patch sizes and the best scaling with smaller patch sizes, leading to the best segmentation performance. We compare the segmentation decoders at patch size 4 in \autoref{tab:decoder_comparison}.

\begin{table}[hb!]
\setlength{\tabcolsep}{0.3em} 
\renewcommand{\arraystretch}{1.1}
\centering
\begin{tabular}{ll||r|r|r}
    & \textbf{Decoder}    & \textbf{Dice} [95\%CI]                & \textbf{Prec.} [95\%CI]           & \textbf{Recall} [95\%CI] \\
    \hline
    \hline
    \multirow{4}{*}{\rotatebox[origin=c]{90}{\textbf{ViTiac-S}}}
        & UNETR      & 53.8 {[}50.5; 57.0{]} & 53.8 {[}50.0; 57.6{]} & \textbf{64.0} {[}60.7; 67.2{]} \\
        & SFPN U-Net & 60.5 {[}57.6; 63.5{]} & 70.1 {[}66.6; 73.6{]} & 59.1 {[}56.0; 62.2{]} \\
        & Upscale    & 59.5 {[}56.5; 62.5{]} & 68.9 {[}65.3; 72.4{]} & 59.5 {[}56.3; 62.6{]} \\
        & MAE        & \textbf{60.6} {[}57.5; 63.7{]} & \textbf{80.5} {[}77.4; 83.6{]} & 53.8 {[}50.5; 57.0{]} \\
    \hline
    \multirow{4}{*}{\rotatebox[origin=c]{90}{\textbf{ViTiac-M}}}
        & UNETR      & 57.4 {[}54.2; 60.6{]} & 60.4 {[}56.7; 64.2{]} & \textbf{64.1} {[}60.8; 67.4{]} \\
        & SFPN U-Net & 57.8 {[}54.7; 60.9{]} & 64.9 {[}61.2; 68.5{]} & 59.4 {[}56.1; 62.7{]} \\
        & Upscale    & 57.1 {[}54.0; 60.1{]} & 74.0 {[}70.5; 77.5{]} & 52.0 {[}48.9; 55.2{]} \\
        & MAE        & \textbf{60.2} {[}57.3; 63.1{]} & \textbf{79.5} {[}76.4; 82.5{]} & 52.6 {[}49.6; 55.6{]} \\
    \hline
    \multirow{4}{*}{\rotatebox[origin=c]{90}{\textbf{ViTiac-L}}}
        & UNETR      & 58.8 {[}55.7; 61.8{]} & 64.1 {[}60.4; 67.9{]} & \textbf{61.8} {[}58.7; 64.8{]} \\
        & SFPN U-Net & 59.6 {[}56.6; 62.6{]} & 67.7 {[}64.0; 71.3{]} & 61.5 {[}58.5; 64.5{]} \\
        & Upscale    & 54.5 {[}51.5; 57.5{]} & 64.6 {[}60.7; 68.4{]} & 55.0 {[}51.8; 58.2{]} \\
        & MAE        & \textbf{61.4} {[}58.6; 64.2{]} & \textbf{83.6} {[}80.8; 86.4{]} & 52.0 {[}49.1; 54.9{]} \\
\end{tabular}
\caption{Mean Dice↑, Precision↑, and Recall↑ with respective 95\% confidence intervals for the (uncalibrated) segmentation decoders. The decoders are added to the same pre-trained encoders with patch size $4^3$.}
\label{tab:decoder_comparison}
\end{table}

\begin{figure}
    \centering
    \includegraphics[width=0.2\textwidth]{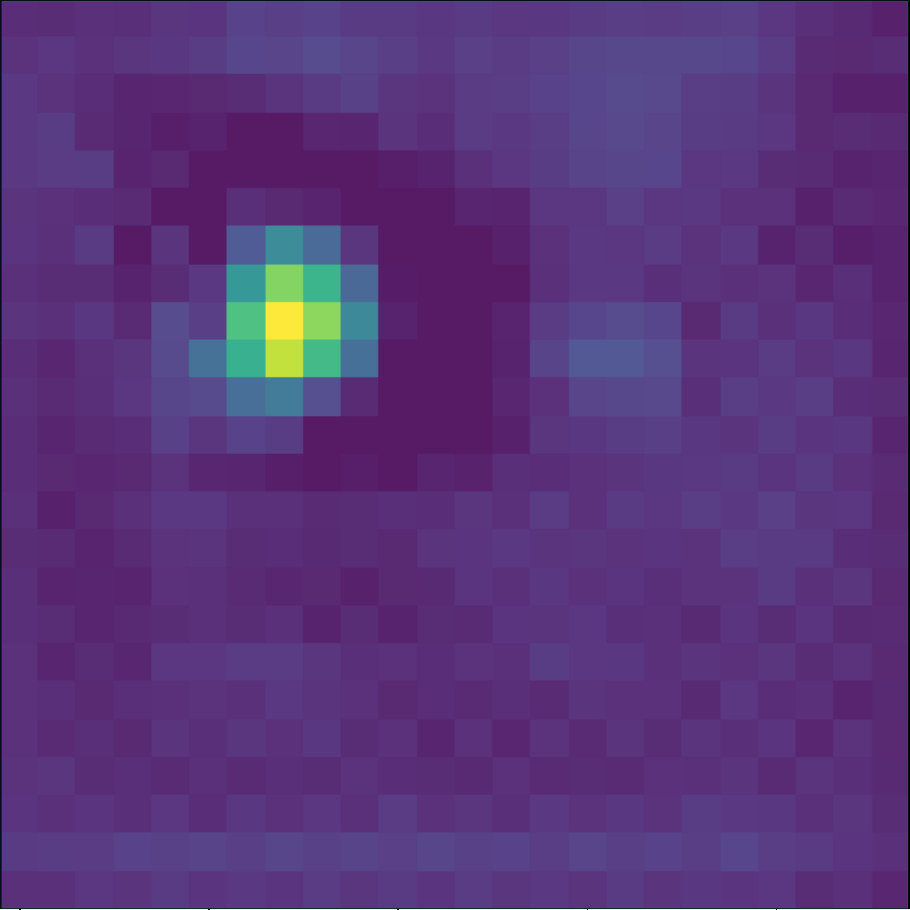}
    \includegraphics[width=0.2\textwidth]{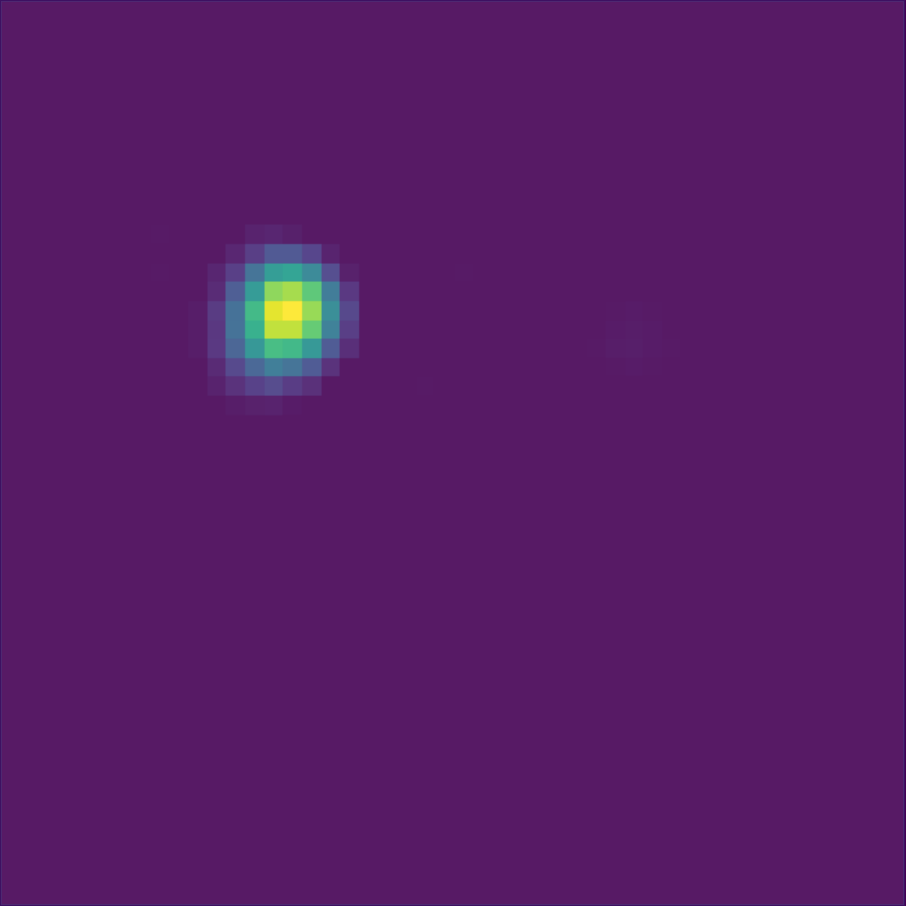}
    \includegraphics[width=0.2\textwidth]{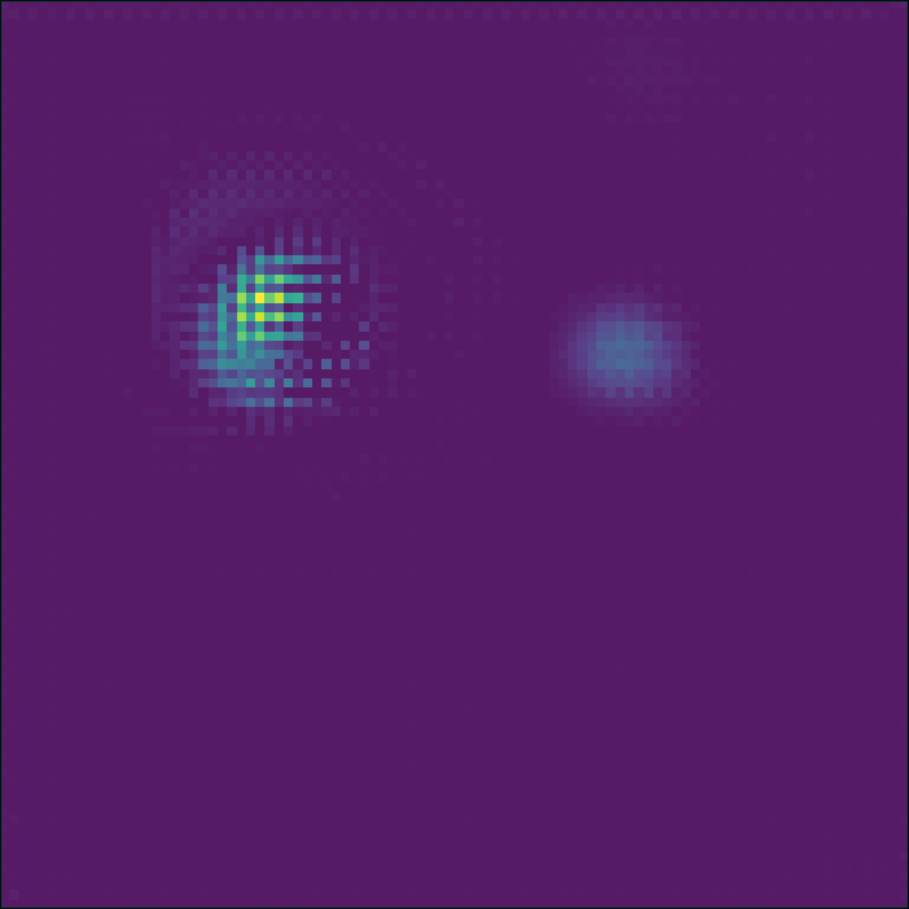}
    \includegraphics[width=0.2\textwidth]{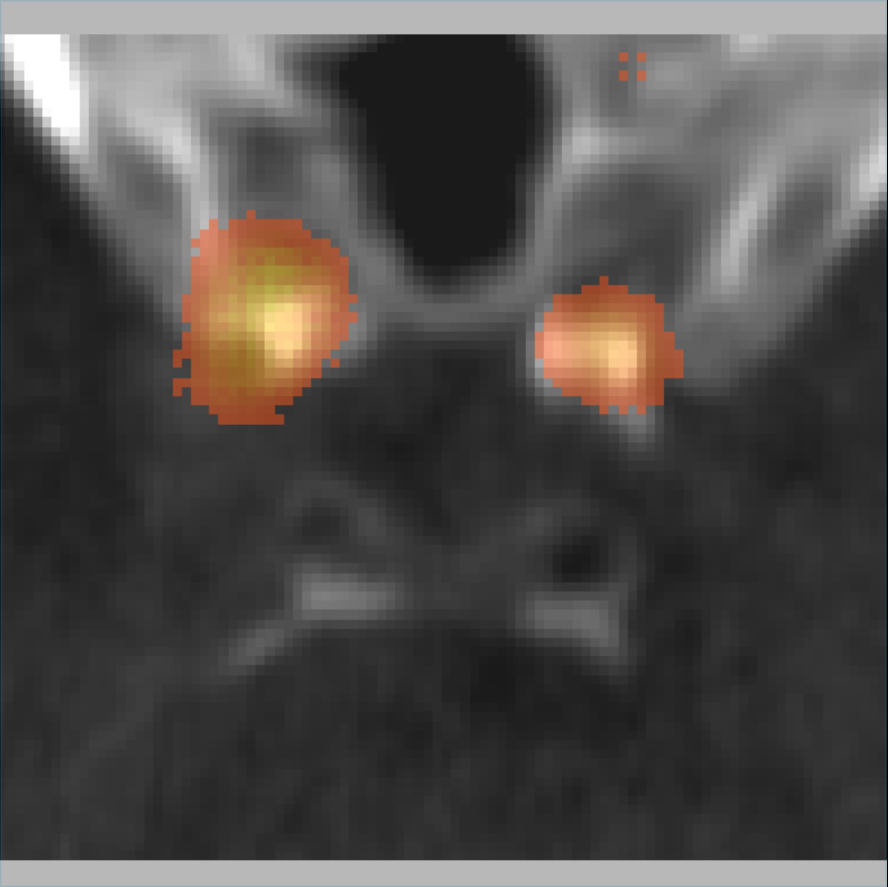}
    \includegraphics[width=0.2\textwidth]{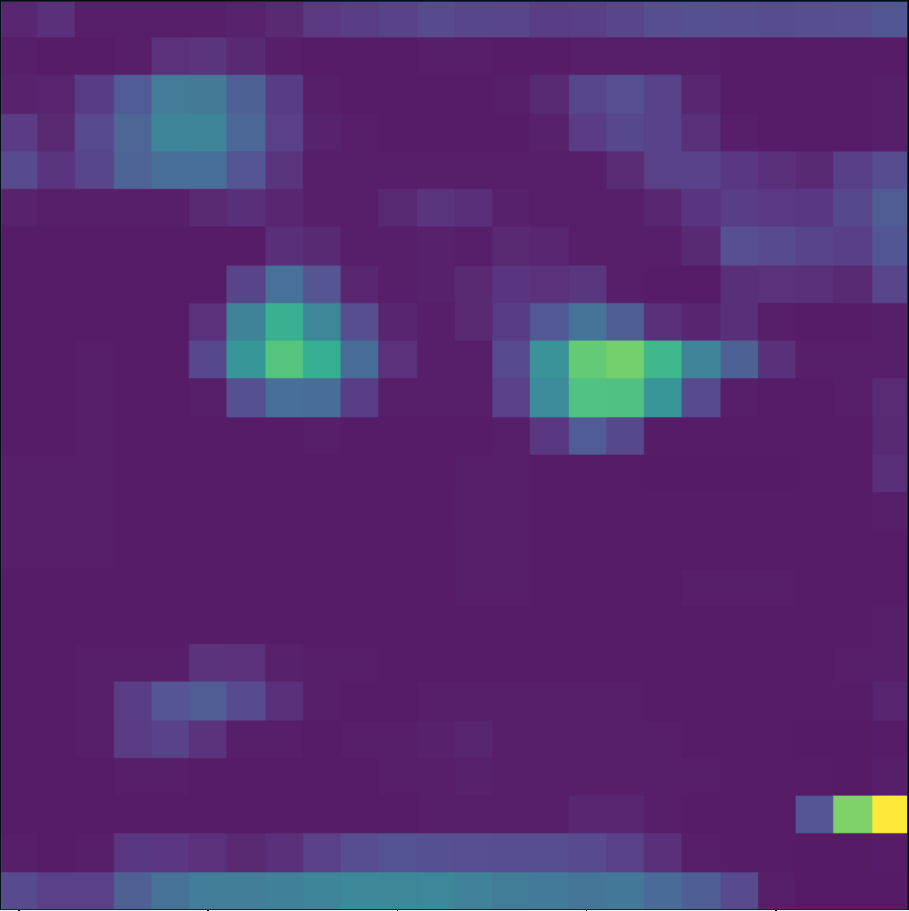}
    \includegraphics[width=0.2\textwidth]{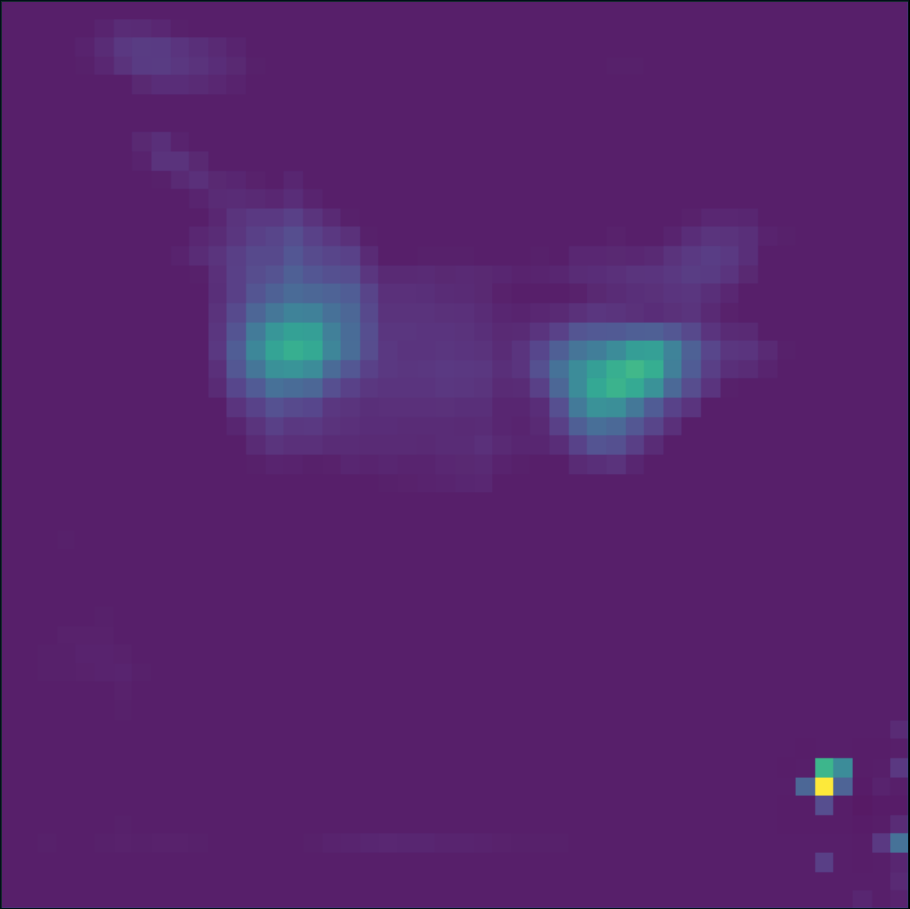}
    \includegraphics[width=0.2\textwidth]{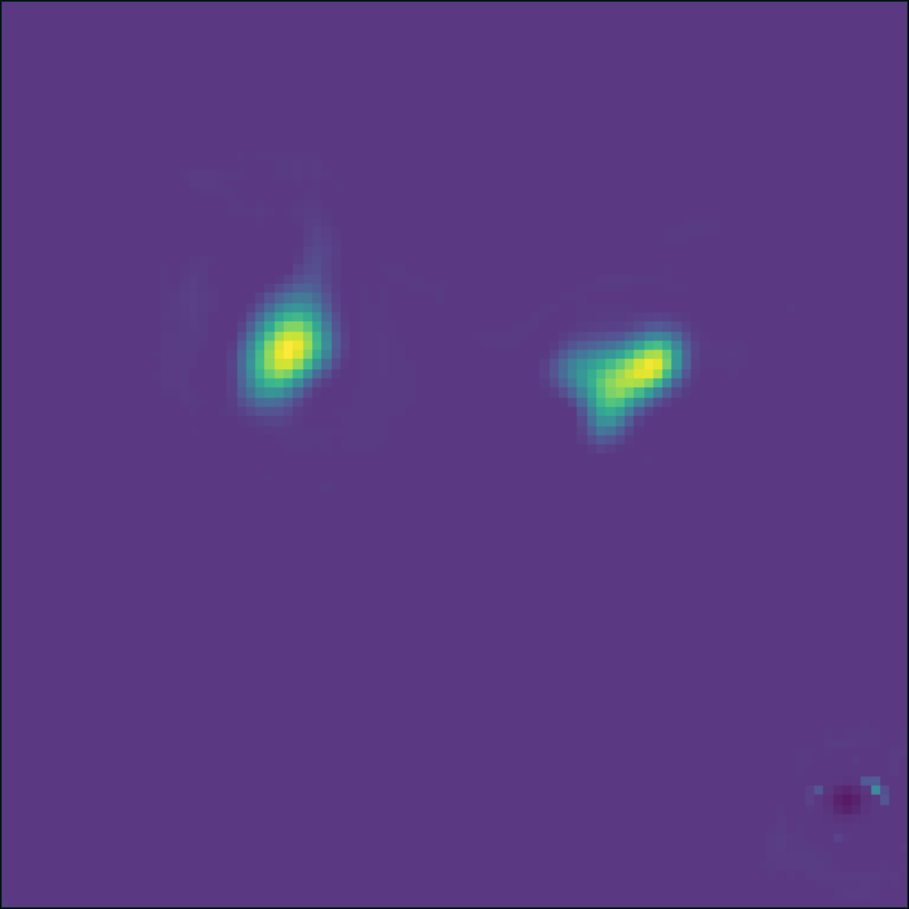}
    \includegraphics[width=0.2\textwidth]{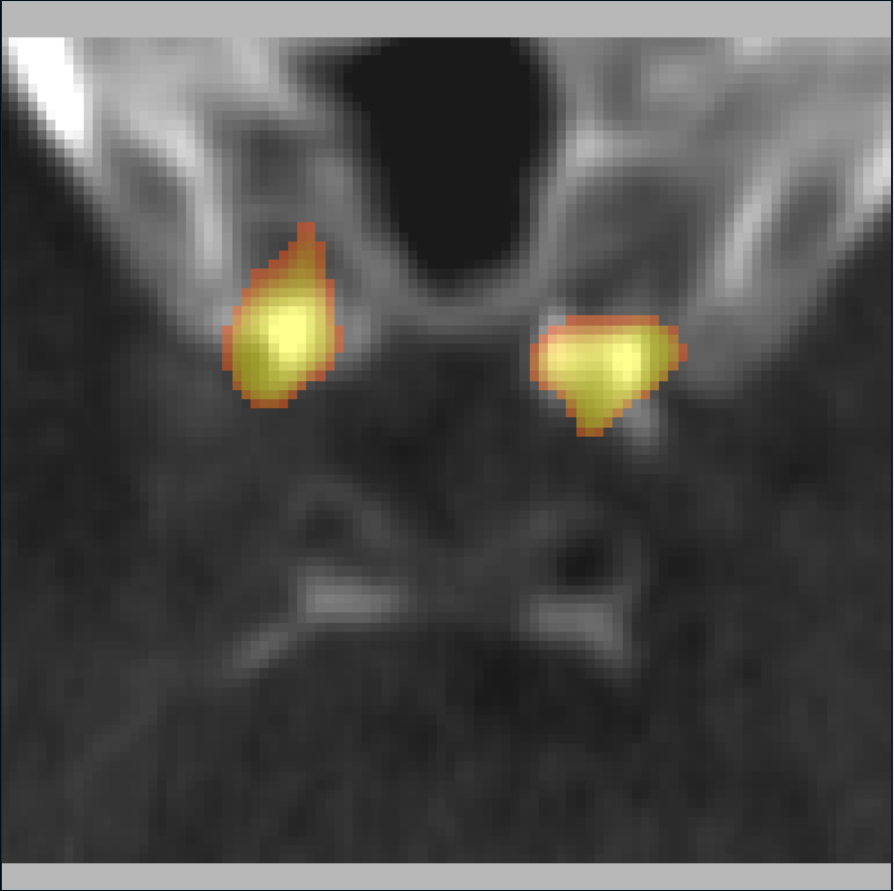}
    \caption{Feature maps left to right from increased scales from the Upscale decoder using transposed convolutions with visible checkerboard artefacts (top) and nearest neighbour interpolation (bottom). Predicted probability masks for the intracranial arterial calcification superimposed on the input (greyscale) slice in the rightmost column.} \label{figure:upscale_feature_maps}
\end{figure}

In our visual inspection of the convolutional decoders' (SFPN U-Net, UNETR, and Upscale decoder) feature maps, we observe signs of checkerboard artefacts, which lead to slightly jagged segmentations in the prediction. We show examples of the effects in the Upscale decoder for patch size $16^3$ in \autoref{figure:upscale_feature_maps}. The average quantitative increase in Dice score using nearest-neighbour interpolation is minor and not entirely consistent.



\subsection{Clinical Relevance}

Accurate evaluation of deep neural networks is crucial for methodological development, but Dice scores are not relevant in clinical practice. For a practical evaluation for the clinical setting, we divide the samples from the test set into four quartiles according to the annotated calcium volume and calculate the correct classification into these groups with our networks' volume predictions. This setting models a future scenario, where patients' risk for neurovascular diseases is assessed based on their individual \acrshort{iac} burden and the associated risk group (i.e. no, low, moderate, or high risk). Based on the risk group, clinicians could advise lifestyle and treatment changes. We find that the calibrated \acrshort{vit}'s segmentations lead to the correct classification in 153 (73\%) cases, while the baseline nnU-Net only correctly classifies 105 (50\%) cases. The calibrated \acrshort{vit} would lead to a 46\% increase in correct risk group classification in the given scenario.



\section{Discussion and Limitations}

Contrary to the natural language and image domains, transformer models have not been able to replace convolutional neural networks as the current state-of-the-art deep learning architecture in 3D medical image segmentation \cite{Isensee24}. Medical imaging datasets are commonly orders of magnitude smaller than natural image datasets \cite{Wald25}, so lower inductive biases can lead to lower segmentation performance when large-scale pre-training is unavailable. Similar to other recent works \cite{Hatamizadeh21,Wald25}, we observe a considerable impact of the \acrshort{vit}'s patch size on the segmentation accuracy. 3D \acrshortpl{vit} compress the input by a much larger factor than 2D \acrshortpl{vit} for the same patch sizes. Originally, we hypothesised that decreasing the patch size for 3D would merely attenuate the compression level, but our experiments showed a relative stability of Dice score around changes in embedding dimensions, with much larger impacts from different patch sizes. We demonstrated that small patch sizes and appropriate pre-training enable transformer models to compete with convolutional neural networks in \acrshort{iac} segmentation.

Most (convolutional \cite{Ronneberger15,Isensee21} and transformer-based \cite{Hatamizadeh21,Wald25}) medical image segmentation models use transposed convolutions. Qualitatively, upscaling with traditional interpolation and regular convolutions, rather than transposed convolutions, results in smoother segmentation boundaries. Beyond checkerboard artefacts, the former offers more flexibility to accommodate a wider variety of patch sizes in the \acrshort{vit}. Thus, we advocate for interpolation with convolution. 

Our proposed calibrated self-supervised \acrshort{vit} improves \acrshort{iac} segmentation in terms of Dice score, robustness against higher slice thicknesses, and simulated clinical utility, compared to the supervised nnU-Net baseline trained on the same routine clinical dataset. On a closer look, the calibrated self-supervised \acrshort{vit} excels at segmenting medium to large \acrshortpl{iac} and slightly struggles with small \acrshortpl{iac}. Future work could explore decreased patch sizes through hierarchical models (as a naïve decrease of patch size increases computational cost quadratically) or hybrid ensembling of transformer and convolutional networks. While previous works report higher segmentation accuracy \cite{Bortsova17}, their models were developed on homogeneous and thin-sliced research-level data. We show the advantage of self-supervised pre-training and the challenges of real-world routine clinical CT imaging for \acrshort{iac} segmentation for the first time, paving the way for large-scale analysis of \acrshort{iac} burden in routinely acquired clinical CT imaging.

Finally, we note an important limitation. The segmentation performance varies substantially based on the properties of the CT imaging. The scans were not only taken on different CT scanners and configurations, but the CT scanning technology evolved over more than ten years of the \acrshort{ist3}. In our analyses, we found slice thickness (resolution of the vertical axis of the 3D volume) particularly impactful on segmentation accuracy.


\section{Conclusion}

We proposed the first self-supervised pre-training for \acrshort{iac} segmentation using heterogeneous routine clinical data. We demonstrated that our calibrated self-supervised models improve a strong supervised nnU-Net baseline by 3.2 Dice points, increase robustness against higher slice thicknesses, and enhance utility in a clinical scenario by achieving 46\% improvement in risk group classification. In ablations of key parameters of the model, we found considerable impact of the patch size of the pre-trained \acrshort{vit}, relative stability around embedding sizes, alternatives to checkerboard artefact-prone transposed convolutions, and the best performance using a transformer segmentation decoder.

Our work aims to facilitate the improvement of accurate automated \acrshort{iac} segmentation from routinely acquired CT head scans, which builds the foundation for large-scale detection and computational description (i.e. quantitative and morphological) of \acrshort{iac}. In the future, we will apply our methods to large, heterogeneous and routinely acquired clinical data with longitudinal outcomes to create a practical risk assessment score for neurovascular diseases. We hope our insights into \acrshort{mae} pre-trained \acrshortpl{vit} for 3D medical image segmentation will be informative beyond our application to intracranial arterial calcification segmentation.

\begin{credits}
\subsubsection{\ackname}
We thank the IST-3 collaborative group for the generous provision of data from the \acrshort{ist3}, all participating patients, and the hospital sites and their supporting staff who were involved in the trial. \acrshort{ist3} was funded from many sources but chiefly the UK Medical Research Council (MRC G0400069 and EME09-800-15) and the UK Stroke Association. The \acrshort{ist3} data is managed and stored within the Systematic Management, Archiving and Reviewing of Trial Images Service (SMARTIS) at the University of Edinburgh. 
B. J. is funded by the Medical Research Council [MR/W006804/1].
G.M. is the Stroke Association Edith Murphy Foundation Senior Clinical Lecturer [SA L-SMP 18\textbackslash1000]. 
J.M.W is part-funded by the UK Dementia Research Institute, which is funded by the UK MRC, Alzheimer's Society and Alzheimer's Research UK, and by the UK National Institute of Health and Care Research. 
M.V.H. is funded by The Row Fogo Charitable Trust [BROD.FID3668413].

\subsubsection{\discintname}
The authors have no competing interests to declare that are relevant to the content of this article. 
\end{credits}

\bibliographystyle{splncs04}
\bibliography{iac_mae_vit}

\end{document}


\section*{Supplementary Materials}

\begin{figure}[hbt!]
    \centering
    \includegraphics[width=0.325\textwidth]{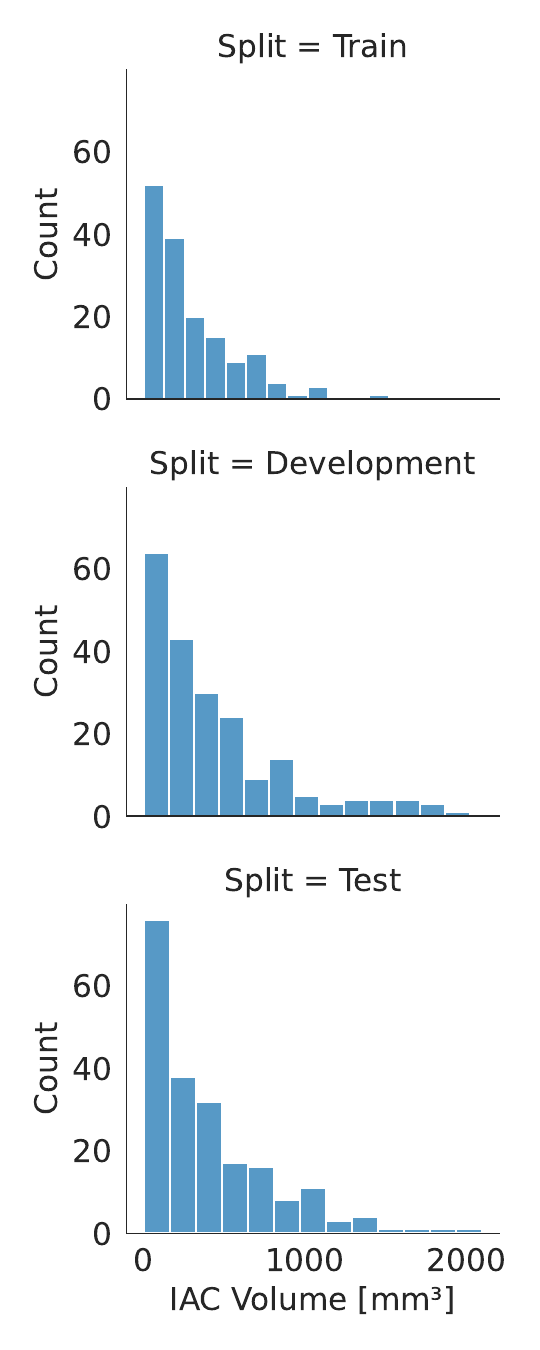}
    \includegraphics[width=0.325\textwidth]{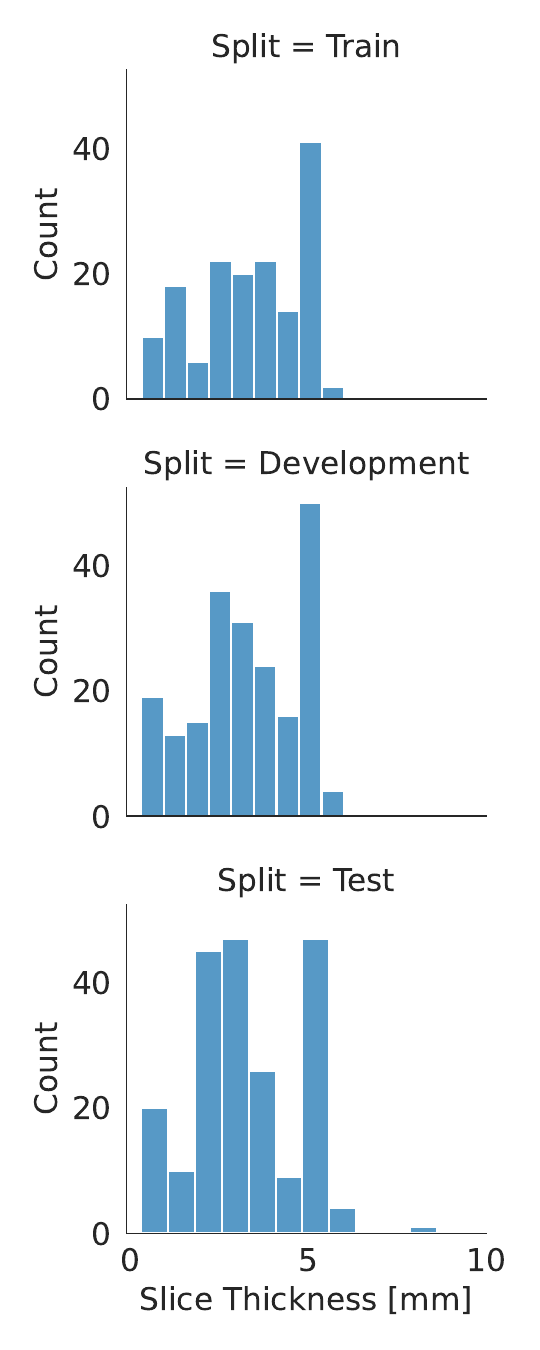}
    \includegraphics[width=0.325\textwidth]{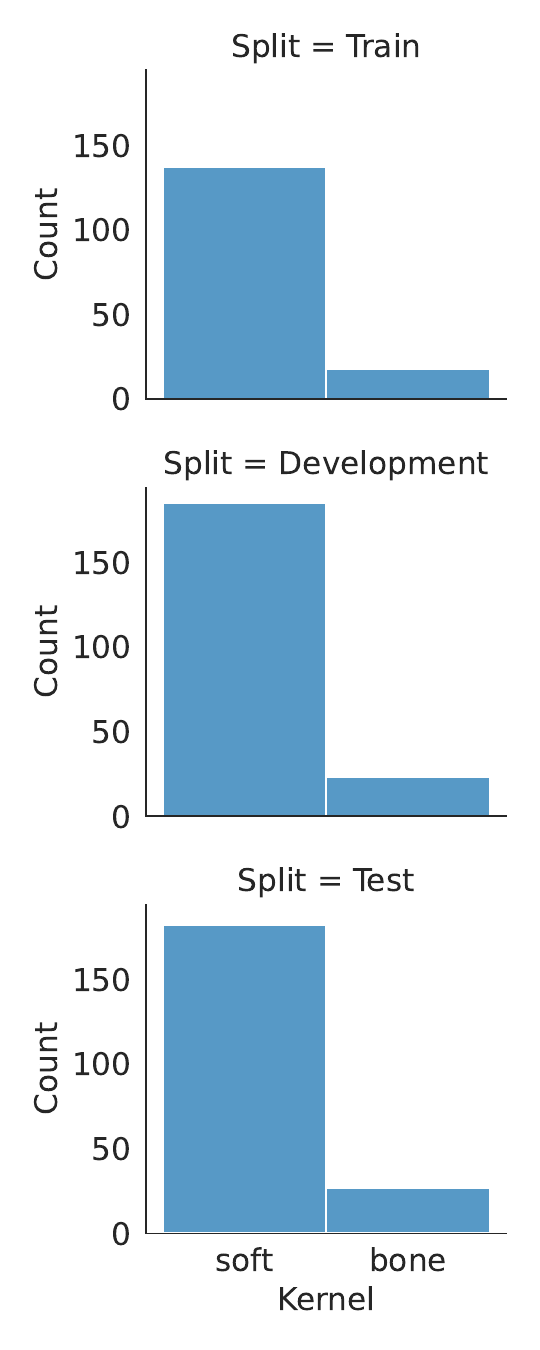}
    
    \caption{Intracranial arterial calcification volume, slice thickness, and convolutional kernel distribution in the annotated subset of the \acrfull{ist3} dataset by their respective dataset splits.}
    \label{suppl:dataset_characteristics}
\end{figure}

\begin{table}[hbt!]
    \centering
    \begin{tabular}{l|rrr|rrr|r}
        \multirow{2}{*}{\textbf{Name}} & \multicolumn{3}{c|}{\textbf{Encoder}} & \multicolumn{3}{c|}{\textbf{Decoder}} & \multirow{2}{*}{\textbf{No. parameters}} \\
            & Embed & Depth & Heads & Embed & Depth & Heads & \\
        \hline
        ViTiac-S & 240 & 12 & 12 & 120 & 8 & 12 & 10M \\
        ViTiac-M & 480 & 12 & 12 & 240 & 8 & 12 & 40M \\
        ViTiac-L & 960 & 12 & 12 & 480 & 8 & 12 & 157M \\
        
    \end{tabular}
    \caption{Vision transformer encoder and decoder configurations for masked autoencoder pre-training. Number of parameters for a patch size of $8^3$, which will differ slightly for other patch sizes.}
    \label{tab:vitiac_configurations}
\end{table}

\begin{figure}[hbt!]
    \centering
    \includegraphics[width=0.8\textwidth]{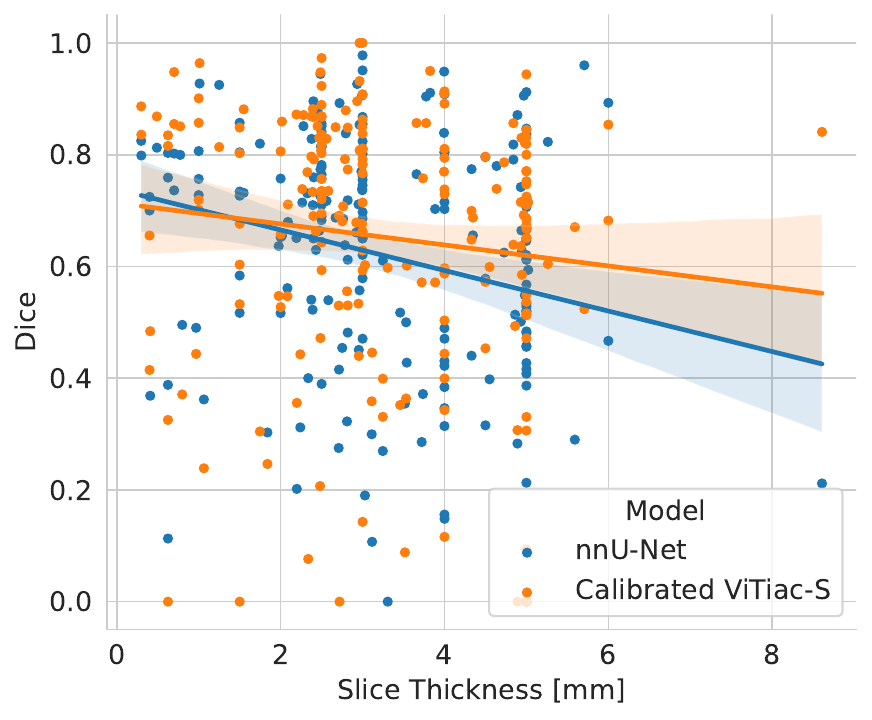}
    \includegraphics[width=0.8\textwidth]{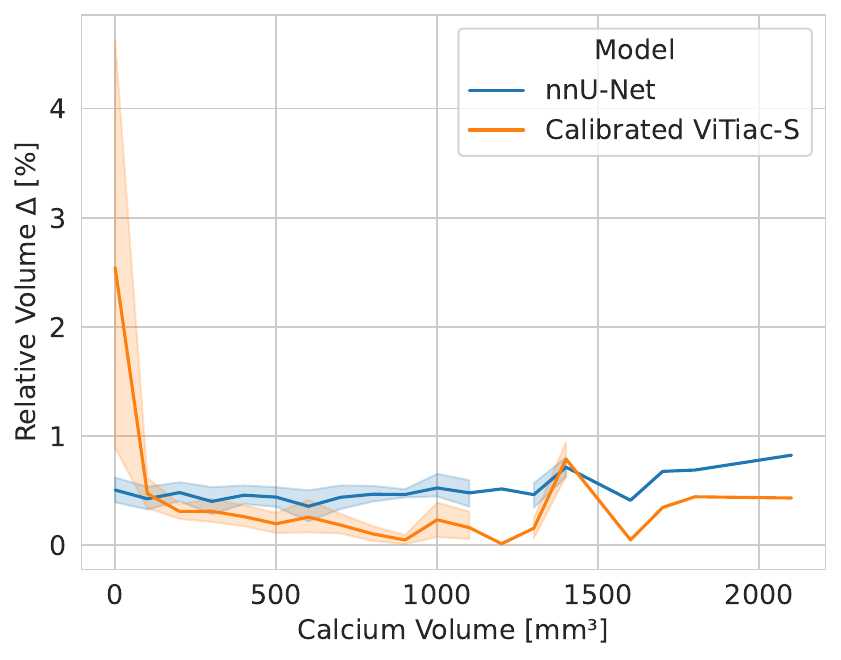}
    
    \caption{Comparison of the Dice score by slice thickness of the CT image series (top) and the relative volume difference (absolute volume difference normalised by the annotated volume) by annotated calcium volume (bottom) between the baseline nnU-Net and the MAE pre-trained and calibrated ViTiac-S model.}
    \label{suppl:performance_characteristics}
\end{figure}